# A System's Approach Taxonomy for User-Centred XAI: A Survey

**Ehsan Emamirad, Pouya Ghiasnezhad Omran, Armin Haller, Shirley Gregor**
Australian National University
*{ehsan.emamirad, p.g.omran, armin.haller, shirley.gregor}*@anu.edu.au

## Abstract

Recent advancements in AI have coincided with ever-increasing efforts in the research community to investigate, classify and evaluate various methods aimed at making AI models explainable. However, most of existing attempts present a method-centric view of eXplainable AI (XAI) which is typically meaningful only for domain experts. There is an apparent lack of a robust qualitative and quantitative performance framework that evaluates the suitability of explanations for different types of users. We survey relevant efforts, and then, propose a unified, inclusive and user-centred taxonomy for XAI based on the principles of General System's Theory, which serves us as a basis for evaluating the appropriateness of XAI approaches for all user types, including both developers and end users.

## 1 Introduction

A recent surge in the development of XAI approaches to tackle the black box nature of advanced Machine Learning (ML) solutions has seen a tremendous effort in literature to study, classify and evaluate XAI methods. Although significantly adding to the ever growing XAI research, existing studies often provide a segregated view of XAI concepts that is dependent on a specific domain and audience. Most of the existing efforts, in addition, do not necessarily capture the requirements of novice users as they intuitively cater to domain experts [Srinivasan and Chander, 2020]. There also appears to be a lack of robust evaluation frameworks and guidelines that both qualitatively and quantitatively measure the appropriateness of the explanations for different types of users [Doshi-Velez and Kim, 2017]. Insights from social and cognitive science in prior works suggest that XAI can only be justified if user satisfaction and trust assessment among other social factors are evaluated [Miller, 2019].

As a response to aforementioned challenges, in this paper, we propose a novel definition for XAI along with a user-centred taxonomy based on the principles of General System's Theory to provide: 1) a full-view framework to move XAI closer to human comprehension, and a qualitative framework that serves as a benchmark to evaluate the suitability of XAI approaches for different types of users.

The remainder of this paper is structured as follows: we first present background in **section 2**, and then provide related work in **section 3**. We discuss the system's view taxonomy of XAI in **section 4**, and in the following **section 5**, we showcase the practical use of the taxonomy by evaluating a select XAI approaches. This paper concludes in **section 6**.

## 2 Background

While the idea of XAI has recently been gaining widespread attention [Holzinger *et al.*, 2018], a universally agreed upon definition of explainability remains as one of the biggest challenges in the study of XAI. Contemporary definitions provide a partial, and sometimes ambiguous, perspective of explainability and what properties it should imply [Ciatto *et al.*, 2020]. This is likely due to the fact that every review paper in literature offers their version of explainability given their specific purpose, domain, and audience [Vilone and Longo, 2020]. Despite the sheer volume of definitions available, they appear to share a limited number of attributes that have been conventionally associated with explanations in the philosophy of science [Pa´ez, 2020; Ciatto *et al.*, 2020]. Some, e.g., [Biran and Cotton, 2017], define XAI as a system's ability to explain the decisions and predictions, and provide transparency of the process through which decisions are made. Others propose XAI as a mechanism to characterise the strengths and weaknesses of the AI agent and convey an understanding of how they will behave in the future [Sanneman and Shah, 2020].

In addition, the majority of existing attempts at XAI approaches are mostly limited to technical articulation of a model's processes [Lyons *et al.*, 2017], which may again only satisfy the explainability needs of domain experts. This is where the concept of explainability is often interchanged with technical transparency and with interpretability in spite of each having a different meaning [Clinciu and Hastie, 2019]. While explainability is arguably about rationale and justification of a recommendation that is understood by end users, transparency refers to understanding of the semantics associated with the computation that goes on in the system [Pal, 2020]. Similarly, interpretability is different from explainability in that it is considered as a formal and logical information about analytical underpinnings of the model which has explanatory features only useful for domain experts [Hoffman *et al.*, 2018]. As such, we posit that

XAI need to be reformulated to account for explainability, transparency and interpretability altogether while ensuring human-comprehensibility of explanations is achieved for all types of users—including both domain experts and end users.

## 3 Related Work

To provide an overview of existing surveys in this field and appropriately compare them, we grouped prior works into five categories and clusters based on distinct aspects of explainability covered by these studies (see Table 1):

1. **XAI approach.** Related surveys relevant to studying and classifying various explainability methods such as whether the explanation is inherent or otherwise the method involves a unique explanator (e.g., surrogate classifier);

2. **ML Type.** Frameworks and architectures that focus on machine learning elements and problems including the type of ML method implemented (e.g., NN), the ML problem (e.g., classification), data type (e.g., unstructured) and whether the ML problem is associated with the features (pre-model), the processes (in-model) or the output (post-model);

3. **Explanation.** Guidelines on the explanation itself, its meaning (e.g., context), format (e.g., textual, image-based), and type (e.g., counterfactual, example-based, causal, justification, etc.);

4. **User domain.** Related models that incorporate user's domain knowledge in developing and retrieving explanations as well as the context, domain, and background of the problem where explanation is often a mandatory requirement (e.g., medical domain); and,

5. **Evaluation.** Studies that introduce and include an evaluative framework for measuring performance of both the ML and XAI's output. These studies differentiate the performance of the original model before and after the explanation is generated.

While all surveys, frameworks, and guidelines presented here contribute to the rapidly growing body of XAI research, to the best of our knowledge, existing surveys generally provide a method-centric framework for classifying and evaluating XAI models—i.e., various concepts and elements surrounding the notion of XAI are typically examined and categorised based on the method itself. Additionally, except two studies (as displayed in Table 1) that highlight the need and requirements of end users, their domain knowledge, background, and context in which they interact with the AI agent to generate explanations, no other survey does provide a unique view and taxonomy of the user-comprehensible explainability. This is consistent with findings in prior studies that confirm the lack of human-in-the-loop requirements in building XAI applications [Chander *et al.*, 2018].

## 4 System's View of XAI

In trying to fill the above gaps, in this survey, we therefore propose a novel and distinct perspective of XAI borrowing key concepts from the Information Systems (IS) research community, and more specifically the General System's Theory [Bertalanffy *et al.*, 1968]. The aims here are twofold: first, to provide a user-centric taxonomy of XAI and its environment in which the user interacts with the AI agent not only as the receiver of explanations but also as a key player in generating them. Second, to provide a unified, holistic, and inclusive framework that would serve as a basis for evaluating the suitability of XAI approaches for all types of users. To demonstrate the viability of our proposed taxonomy, we conducted a comprehensive meta-analysis of a wide array of XAI approaches, and classified a series of inter-related elements contributing to the design of XAI methods, which emerged in our review. These elements include the concept of explanation, types of explainability, properties of explainability, application domain, human-in-the-loop, XAI approaches and methods, and performance and evaluation.

By bringing these elements together, we can conceptualise them based on their interrelationship and underlying contribution in designing XAI architectures. Figure 1 illustrates the inter-dependency among the various XAI elements with a respective notional meta model. In the following, given a problem domain, a typical high-performing black-box ML method, such as Deep Neural Networks [Gunning and Aha, 2019], would require an explanation that is typically generated through a specific XAI approach or method. The required explanation should include key types and properties that are easily comprehended by the user. The end user also evaluates the performance of the original ML model as well as the XAI model to decide whether the explanation is comprehensible. A user's type and background knowledge, the context in which the explanation is required, the user's involvement and interaction with the AI agent in producing or making sense of the the explanation along with the effort exerted to generate those explanations would additionally play a critical part in establishing whether the XAI outcome is suitable, and hence, trustworthy [Gregor and Yu, 2002]. Researchers have shown that trust further serves as a mandatory requirement for wide acceptance of AI applications, particularly in scenarios such as medical domain that requires absolute justification and reliability of predictive results [Holzinger *et al.*, 2018].

Applying the principles of General System's Theory, we can further infer that the concepts in the meta model are a set of related and inter-related elements for which there is sufficient coherence and interactivity to make them as a whole useful and meaningful. Here, this holistic view necessitates to consider XAI as a system of parts in an interaction, instead of a standalone concept, approach or method. As such, this study further posits a fresh definition of XAI as follows:

**Definition 1.** *Given a particular application domain, XAI is a system of elements, with supporting methods and behaviours to ensure AI agents, including the generated outputs, are understood and trusted by relevant users.*

Apart from a newly defined perspective for XAI, containing its concepts within a System's view taxonomy would be unique in a number of different ways. Firstly, the existing XAI approaches and methods can be further classified based on the Input-Process-Output (IPO) model, which is a widely used approach in IS theory and practice. The pertinent opportunity here is to investigate whether an ML method could be made explainable at the input level (pre-model), process level (in-model) or at the output level (post-model). This way of classification is similar to understanding whether a model is inherently explainable (process level explainability) or requires post hoc interpretability technique (output level explainability). Secondly, this perspective offers a user-centred approach where users of all types, irrespective of their background knowledge, are engaged and interacted with throughout the development, understanding and trusting of AI agents. Thirdly, the nature of explanations in conjunction with its type and property collectively make up the notion of explainability. Additionally, explainability performance is considered as equally important as the original model's performance when establishing whether an explanation is appropriate. Finally, a qualitative benchmark can be developed to help decide whether an XAI method is user-comprehensible, or otherwise, to identify the shortcomings that need to be addressed to make the method user comprehensible. In the following, we present a brief review of each element provided above.

| Study/Category | Intrinsic/Post-hoc | Global/local | Model specific/agnostic | Explanator | Black box type | Pre-/in-/post-model | ML problem | Data type | Explanation properties | Explainability need | Explanation dynamics | Explanation type | Explanation format | User knowledge | Application scenarios | Model performance |
|---|---|---|---|---|---|---|---|---|---|---|---|---|---|---|---|---|
| | **Explainability technique** | | | | **ML type** | | | | **Explanation** | | | | | **User domain** | | **Evaluation** |
| [Adadi and Berrada, 2018] | ✓ | ✓ | ✓ | | | | | | | | | | | | | |
| [Anjomshoae et al., 2019] | | | | ✓ | | | | | | ✓ | ✓ | ✓ | | | ✓ | ✓ |
| [Arrieta et al., 2020] | ✓ | | ✓ | | ✓ | | | | ✓ | | | ✓ | | | | |
| [Carvalho et al., 2019] | ✓ | ✓ | ✓ | ✓ | | ✓ | | | ✓ | ✓ | | ✓ | | | | |
| [Chakraborti et al., 2020] | | ✓ | | | | | | | ✓ | | | | | | | ✓ |
| [Das and Rad, 2020] | ✓ | ✓ | | | | ✓ | ✓ | ✓ | | | | | | | | ✓ |
| [Došilović et al., 2018] | ✓ | | | | | | | | | | | | | | | |
| [Du et al., 2020] | ✓ | ✓ | | | | | | | | | | | | | | |
| [Gilpin et al., 2018] | | | | ✓ | | | | | ✓ | | ✓ | | | | | |
| [Guidotti et al., 2019b] | ✓ | ✓ | ✓ | ✓ | ✓ | | ✓ | ✓ | ✓ | | | ✓ | | | | |
| [Li et al., 2022] | ✓ | | | | ✓ | | | ✓ | | | | ✓ | | | | |
| [Mohseni et al., 2021] | ✓ | ✓ | ✓ | | | | | | ✓ | | | ✓ | | ✓ | ✓ | ✓ |
| [Nunes and Jannach, 2017] | | | | ✓ | ✓ | | ✓ | ✓ | ✓ | ✓ | | | ✓ | ✓ | ✓ | |
| [Zhang and Chen, 2020] | | | | ✓ | ✓ | | ✓ | ✓ | ✓ | | | ✓ | | | | ✓ |
| [Vilone and Longo, 2020] | ✓ | ✓ | ✓ | ✓ | ✓ | | ✓ | ✓ | ✓ | | | ✓ | ✓ | | | ✓ |

Table 1: Examined related surveys.

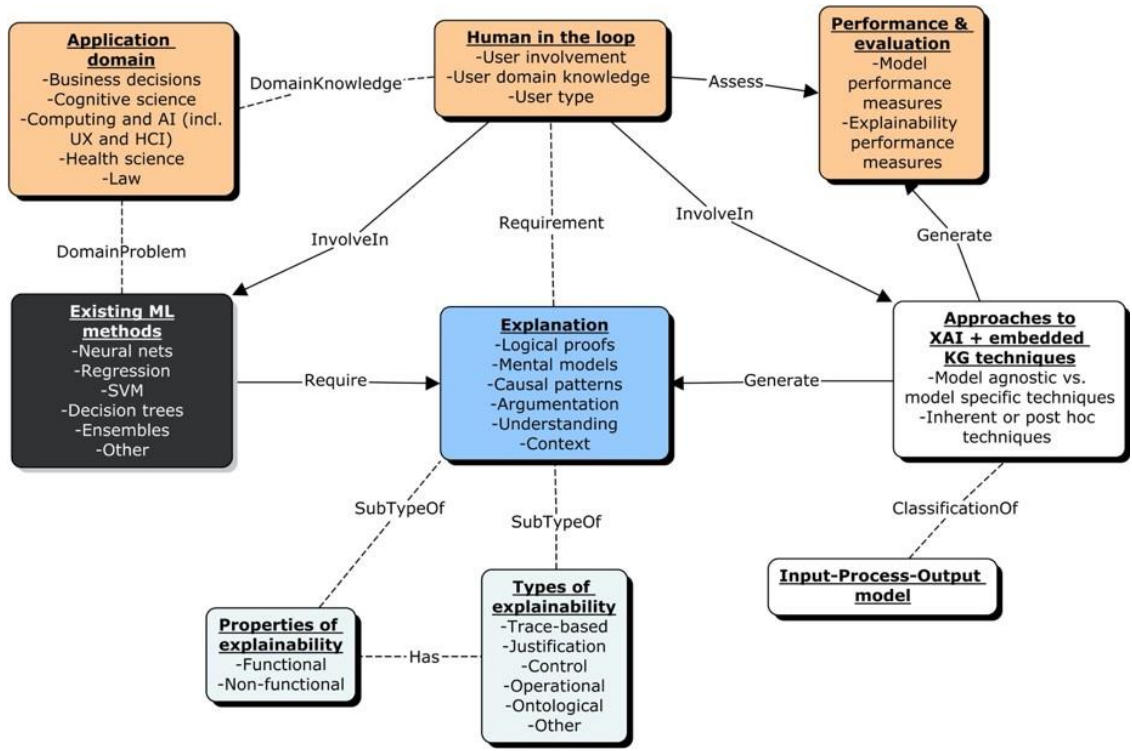

Figure 1: Proposed XAI meta model.

### 4.1 Explaining Explanations

Understanding and defining explanation is a theoretical and practical challenge due to its contextual nature and different meaning in every domain [Craik, 1952]. The study of explanation, while it has a long tradition in philosophy and is mostly related to cognitive and intuitive theories, offers important new perspectives in AI [Fagin *et al.*, 2004], particularly, in terms of establishing whether the explanations that humans expect from systems match those that they typically do from other humans [Gregor and Yu, 2002].

The literature on explanation is as diverse as it is massive with large volumes of research devoted to deductive proofs, patterns of causation, explanatory and mental stances, argumentation, and understanding. In the contemporary works concerning explanations, a set of basic laws are stated as axioms where deductive proof of those laws are considered as explanation [Keil, 2006]. Explanations are also considered as tracing the line of reasoning when investigating causalities. Here, explanations are logical occurrences originating from facts from certain events [Von Wright, 1971]. Further, explanations are thought of as mental models, which can vary from formal representations of logical patterns to image-like internal representations of a system [Norman *et al.*, 1983]. Regardless of the nature of explanations, they complement understanding by forming the basis of arguments, generalizations and constructing interpretations [Wiggins *et al.*, 2005].

### 4.2 Types of Explainability

Like the explanation itself, much of the recent literature has been devoted to discussing the classification of explanations. Many studies propose that explanation should answer several questions to be deemed exhaustive [Vilone and Longo, 2020], primarily including trace-based questions [Gregor and Yu, 2002; Preece, 2018]. The suitability of explanation type is also determined based on its content, format, the time it is expected to be produced, and the user involvement in deciding the explanation type.

The majority of explanation types reported in literature are arguably limited to the most common queries of 'how' and 'why' [Mohseni *et al.*, 2021]. For example, contrastive and counter-factual explanations are about providing justification, argument or counter argument for the produced explanation [Hoffman *et al.*, 2018]. In another attempt to address the 'why' queries, mechanistic and causal explanations are provided which, in essence, relate to trace-based explanations discussed in numerous studies [Abdul *et al.*, 2018]. There are also case-based, contextual and example-based explanations that relate to the context of explanation as well as similar cases or examples that will help establish the 'how' aspect of the explanation [Cai *et al.*, 2019].

### 4.3 Properties of Explainability

The concept of explainability also sits at the intersection of several properties that are often dependent on the features of the ML model and the user ([Doran *et al.*, 2017]). In reviewing literature, some of these properties could be considered functional, for example, fidelity which is a measure corresponding to the reliability and accuracy of approximation of the black box behaviour through the explanation [Pedreschi *et al.*, 2019], while others are regarded as non-functional such as transparency, understanding, reasoning, and trustworthiness among others [Nunes and Jannach, 2017]. In both instances, it is argued that an AI agent is not explainable if it does not support both properties [Lipton, 2018; Hoffman *et al.*, 2018; Hagras, 2018; Doran *et al.*, 2017].

### 4.4 Application Domain

Generating explanations involves careful consideration of the application domain and context [Rudin, 2019], and within that the nature of the user and their role in producing explanations. Explanations should be generated based on the match between the complexity of the explanation and the complexity of human's capability and domain knowledge [Collaris and van Wijk, 2020]. Explanations should also be tailored upon the role of the human interacting with the AI agent and the user's involvement in producing them [Miller, 2019].

Numerous application domains explored in detail in literature. Of these include explanations in scenarios such as *cognitive science*, which is mostly described as an interactive process of knowledge transfer [Srinivasan and Chander, 2020; Miller, 2019; Floridi *et al.*, 2020; Gregor and Yu, 2002]; *computer science*, which is about analytical and mathematical transparency of the model [Lyons *et al.*, 2017; Hoffman *et al.*, 2018]; *human computer interaction*, which is mostly associated with user perception and comprehension of the systems [Sanneman and Shah, 2020; Pa´ez, 2020; Miller, 2019; Hagras, 2018]; and highly consequential domains of *health science* and *law*, which trustworthiness, causality, informativeness, fairness, accessibility and privacy awareness are said to be critical factors [Biran and Cotton, 2017; Arrieta *et al.*, 2020; Felzmann *et al.*, 2019].

### 4.5 Human-in-the-Loop

Explainability in ML models in large part depends on their intended users, their type and background knowledge, their involvement in both training of the ML model and also processing the explanations, their effort in exerting to access explanations, as well as their ability to comprehend the outputs based on their cognitive behavioural capability, limitations and domain knowledge [Gregor and Yu, 2002; Ribera and Lapedriza, 2019]. Development of a user comprehensible XAI requires an understanding of the requirements of the human-in-the loop seeking explanation and their engagement early on in the process of designing explainable models—a key condition distinctively lacking in existing XAI efforts [Srinivasan and Chander, 2020].

In terms of the user type, there are a number of explainability requirements set out for different users which should be considered and differentiated when designing explainable models. For example, explainability requirement for domain experts or AI *developers* can be described as an understanding of whether the system is working properly to identify and remove bugs while the same requirement for *end users*, or novice users who receives those explanations, is described, on the contrary, as a sense of reasoning and justification of a system's behaviour to build trust [Felzmann *et al.*, 2019]. The end user is, in addition, responsible for making sense of the decisions recommended by an AI agent through processing, compiling and comprehending its outputs with a great degree of correspondence to their cognitive "intuition" [Doran *et al.*, 2017]. In order for the actions of an agent to be comprehensible, it is important to also consider the specific context, background knowledge, and interests and limitations of end users [Hagras, 2018]—end user ultimately has the best possible understanding of the agent and the predictive reliability of its output [Pa´ez, 2020].

### 4.6 XAI Approaches and Methods

There are growing number of approaches aimed at effectively realising XAI, see e.g., [Guidotti *et al.*, 2019b; Gilpin *et al.*, 2018; Miller, 2019; Adadi and Berrada, 2018; Vilone and Longo, 2020; Zhang *et al.*, 2022]. Existing XAI approaches can be divided into two main categories of inherently interpretable methods that are explainable by design, e.g., symbolic AI, decision trees, regression and etc., and post hoc interpretability methods that are designed to extract understandable information from a non-transparent black box model, e.g., surrogate models, feature extraction methods, deep explanation among many others.

Of the existing methods, they can further be classified based on whether they provide explanations associated with individual instances or the entire predictive model. The local instance interpretations, also known an as model-specific methods, provide a local approximation only pertaining to an individual prediction. Whereas in terms of global model interpretations, or model-agnostic methods, they provide approximations of the black box model aiming to offer global replicability and understanding of the model in its entirety. Finally, XAI approaches can also be categorised according to the user involvement in generating the explanations. Static methods simply present explanations to the end user while interactive methods provide users with the opportunity to interact with explanations through output formats such as visuals, conversations, rules, and to name a few.

### 4.7 Performance and Evaluation

One of the main goals of XAI is to improve performance, learning and user perception of AI agents [Gregor and Yu, 2002]. However, there are often trade-offs between explainability and performance of the agent. Although these trade-offs have been debated in literature, e.g., see [Rudin, 2019], the common belief is that as internal complexity of an AI agent increases, the model seems to be less explainable [Hagras, 2018]. The task ahead is therefore to design and develop XAI methods that are able to produce explainability while maintaining high levels of prediction accuracy [Holzinger *et al.*, 2017], in addition to striking an appropriate balance between information completeness and user comprehensibility [Pa´ez, 2020]. A further challenge is to institute whether the explanation works and whether the end user has developed an accepted level of understanding of and trust in the AI agent. In this case, it is important to design more specific and broader qualitative measures to assess the goodness of explanation, explanation satisfaction, the trust in the system as well as other social factors as appropriate [Hoffman *et al.*, 2018; Miller, 2019].

## 5 Evaluation of XAI Approaches

As observed earlier, existing frameworks are not necessarily designed as a tool to evaluate the performance of XAI approaches, particularly, in terms of their user comprehensibility. Inspired by the above discussions, here, we showcase the practical use of our proposed taxonomy applied to a select range of XAI approaches available in literature. Summarized in Table 2, these methods have been selected based on a number of factors: 1) applying the taxonomy to a variety of methods and problem domains given their user considerations, 2) reviewing and classifying a mix of general focused methods, such as surrogate classifiers, and narrow focused methods such as deep tensor, and 3) credibility and relevance of these methods reported in literature. Evaluating these methods draw several key implications:

**User considerations.** All methods appear to require users to possess background knowledge equivalent to those of domain experts. It is, in turn, unclear whether the explanation generated from these methods could yield properties that are easily comprehended by novice users. This is further reinforced by the fact that majority of XAI methods have computing as their application domain. Another key observation here is that except knowledge-based methods and LIME, intended user type in the most of existing efforts is domain experts. This is consistent with the requirement that specific users, such as domain experts, be involved in producing explanations with no explicit indication as to whether such explanations meet the requirements of novice users.

**Explanation.** Most of reviewed methods are model agnostic aiming at providing explanations for individual predictions. Similarly, majority of methods provide explanations at an output level through post hoc interpretability techniques such as deploying surrogate classifiers while a handful focus on providing transparency of the internal workings of the ML model through techniques such as feature importance extraction. Here, there would be opportunities to investigate

| Selected XAI Approach | | Classifier type | Domain | Agnostic/Specific | Global/Local | Explanation type | Functional property | Transparency | Understanding | Reasoning | Trustworthiness | User knowledge | User type | User involvement | Performance |
|---|---|---|---|---|---|---|---|---|---|---|---|---|---|---|---|
| | | | | | | | | \multicolumn{4}{c}{Non-functional} | | | | |
| [Angelov and Soares, 2020] | xDNN | PR | M | Agn | Loc | R | | ✓ | ✓ | | | H | E | | |
| [Anjomshoae et al., 2019] | CI&CU | MC | B | Agn | Loc | J, C | | ✓ | ✓ | ✓ | | H | E | ✓ | |
| [Casalicchio et al., 2018] | SFIMP | FI | C | Agn | Both | R | | | ✓ | ✓ | | H | E | ✓ | |
| [Shrikumar et al., 2017] | DeepLIFT | FI | C | Agn | Loc | R | | | ✓ | | | H | E | ✓ | ✓ |
| [Guidotti et al., 2019a] | LORE | SM | C | Agn | Loc | R, C | ✓ | | ✓ | ✓ | | M | E | ✓ | |
| [Collaris and van Wijk, 2020] | XExplore | SM | C | Agn | Both | R | ✓ | | ✓ | ✓ | | H | E | ✓ | ✓ |
| [Ribeiro et al., 2016] | LIME | SM | C | Agn | Loc | R | ✓ | | ✓ | | ✓ | M | N | ✓ | ✓ |
| [Ehsan et al., 2019] | ARM | TS | C | Agn | Both | R, J | | ✓ | ✓ | ✓ | | M | E | | |
| [Lou et al., 2013] | GA2M | – | C | Sp | Loc | R | | | ✓ | | | H | E | ✓ | |
| [Ai et al., 2018] | – | KB | C | Agn | Both | R | ✓ | | ✓ | ✓ | | L | N | | ✓ |
| [Fuji et al., 2019] | Deep Tensor | KB | M | Agn | Both | R, J | ✓ | | ✓ | ✓ | ✓ | M | N | ✓ | ✓ |

Table 2: Selected XAI approaches. *PR* = Prototype-Based, *MC* = Multi Decision Criteria, *FI* = Feature Importance, *SM* = Surrogate Model, *TS* = Training Explanation Set, *KB* = Knowledge-Based | *B* = Business Applications, *M* = Medical Science, *C* = Computer Science | *NN* = Neural Networks, *LRM* = Linear Regression Model, *SVM* = Support Vector Machines, *RFC* = Random Forest Classifier, *GLM* = General Linear Model, *NB* = Naïve Bayes, *DT* = Decision Tree, *ACLF* = All Classifiers *CF* = Collaborative Filtering | *Agn* = Agnostic, *Sp* = Specific | *Loc* = Local | *R* = Reasoning, *J* = Justification, *C* = Contrastive | *H* = High, *M* = Medium, *L* = Low | *E* = Expert, *N* = Novice

approaches to further incorporate input-level interpretability into the XAI method. In all the methods investigated, along with identifying and explaining the causal pattern of predictive decisions, line of reasoning for predictive behaviours seem to be the two main types of explanations. In terms of the property of explanation, while only a few methods discuss fidelity and faithfulness as prerequisite functional properties, understanding has been shown to be the most important non-functional property among all methods.

**Performance.** Some of the methods provide a measure or a collection of evaluative indicators to show improved performance with respect to the original model's performance before and after explanation is generated in terms of higher prediction accuracy, shorter training time, lower computation resource, lower variation in results, and higher precision and recall. None of the methods, however, discuss the trade-off between model performance and explainability performance, particularly, in the context of assessing the appropriateness and comprehensibility of the explanation from an end user's point of view. As before, knowledge-based methods along with LIME approach are the only methods that provide performance metrics that, to some degree, meet the explainability requirements of end users.

## 6 Conclusions

Moving forward, it would be essential to develop evaluative strategies that quantify user behaviours engaging with the XAI method before and after explanations are provided. The aim here is to not only measure, both qualitatively and quantitatively, the explanation performance, but also establish, from an end user's perspective, whether the user has pragmatic understanding of and trust in he AI agent. Additionally, building domain-independent XAI architectures that are integrated with knowledge-based systems would be a prominent paradigm that deserves further investigation. The goal here is to enhance user comprehensible explanations for different types of users by embedding knowledge-based systems into existing AI models.

In this paper, we surveyed a range of XAI studies and methods, and provided a detailed meta review of various elements and concepts surrounding the notion of XAI. We also proposed a novel perspective of XAI based on the System's Theory along with a full-view, and user centred taxonomy for XAI. In the end, we demonstrated the viability of the taxonomy by applying it to a select few XAI approaches available in the literature to qualitatively evaluate their suitability for users regardless of their domain knowledge.

## References

[Abdul et al., 2018] Ashraf M. Abdul, Jo Vermeulen, Danding Wang, Brian Y. Lim, and Mohan S. Kankanhalli. Trends and trajectories for explainable, accountable and intelligible systems: An HCI research agenda. In *CHI*, page 582, 2018.

[Adadi and Berrada, 2018] Amina Adadi and Mohammed Berrada. Peeking inside the black-box: A survey on explainable artificial intelligence (XAI). *IEEE Access*, 6:52138–52160, 2018.

[Ai et al., 2018] Qingyao Ai, Vahid Azizi, Xu Chen, and Yongfeng Zhang. Learning heterogeneous knowledge base embeddings for explainable recommendation. *Algorithms*, 11(9):137, 2018.